\documentclass[11pt]{article}
\usepackage[margin=1in]{geometry}
\usepackage{graphicx}
\usepackage{float} 
\usepackage{amsmath,amssymb,amsthm}
\usepackage{booktabs,longtable,array,tabularx}
\usepackage{xcolor}
\usepackage{hyperref}
\usepackage[numbers]{natbib}

\title{Causal-based approaches to explain and learn from self-extension - A review}
\usepackage{authblk}
\author[1]{Rebeca Marfil}
\author[2]{Pablo Bustos}
\author[1]{Antonio Bandera}
\affil[1]{Universidad de Málaga, Spain}
\affil[2]{Universidad de Extremadura, Spain}
\date{}

\begin{document}
\maketitle

\begin{abstract}
The last decades have seen a revolution in autonomous robotics. Deep learning approaches and their hardware implementations have made it possible to endow robots with extraordinary perceptual capabilities. In addition, they can also benefit from advances in Automated Planning, allowing them to autonomously solve complex tasks. However, on many occasions, the robot still acts without internalising and understanding the reasons behind a perception or an action, beyond an immediate response to a current state of the context. This gap results in limitations that affect its performance, reliability and trustworthiness. Deep learning alone cannot bridge this gap, because the reasons behind behaviour, when it emanates from a model in which the world is a black-box, are not accessible. What is really needed is an underlying architecture based on deeper reasoning. Among other issues, this architecture should enable the robot to generate explanations, allowing people to know why the robot is performing, or has performed, a certain action, or the reasons that may have caused a certain plan failure or perceptual anomaly. Furthermore, when these explanations arise from a cognitive process and are shared, and thus validated, with people, the robot should be able to incorporate these explanations into its knowledge base, and thus use this understanding to improve future behaviour. Our article looks at recent advances in the development of self-aware, self-evolving robots. These robots are designed to provide the necessary explanations to their human counterparts, thereby enhancing their functional capabilities, in the quest to gain their trust.
\end{abstract}


\section{Introduction} \label{sec-intro}

The upcoming Society 5.0 represents a new paradigm in which people and artificial beings cooperate in daily life routines, environments, and interactions \cite{gladden:2019}. This cooperation is intended to be natural and intuitive, and the new artificial actors are expected to behave appropriately. But the concept of what is an 'appropriate' behaviour is context-specific and just as complex and diverse as societies and individuals are. Moreover, it depends not only on functionality but also on acceptability, accessibility, and adaptability \cite{mataric:2017,sgorbissa:2018}. Service robots constitute one of the technologies with the highest number of potential applications and are also one of the most deeply affected by these technical and contextual complexities \cite{seibt:2020}. Service robots, being inevitably social when used in daily life scenarios, face the most challenging technical, ethical, social, and legislative demands \cite{tian:2021}.

Socially-aware Robotics (SAWR) is an emergent area of research that seeks to understand how cognitive robots can become aware of their social context and use this capability to behave as more accessible, accepted, and useful devices, being able to establish more adequate and effective interactions to help humans \cite{rossi:2020}. A robot aiming to exploit autonomous capabilities that are socially enhanced requires sensing its environment and achieving a certain level of understanding about what happens within this environment. But to be truly socially aware, a robot must not only purposely react to perceived changes: it should also predict the incoming events and anticipate their consequences, selecting the best possible and most understandable action and respecting social conventions \cite{sandini:2020,hellstrom:2018}. And their behaviour must also be correctly interpreted by the people with whom they share their environment. The process of anticipation reduces the response time and has the potential to be perceived by humans as more natural, increasing the chances for the robot to produce a socially adequate type of behaviour. These robot-environment dynamics anchor the robot to its world and open the way to online autonomous learning. Thus, socially-aware robots are defined here as proactive machines that predict, anticipate, evaluate and learn from their social context to adapt their behaviour to that context. Ultimately, a socially-aware robot must be perceived as trustworthy by humans, a condition that requires transparency, self-explication and legibility in its behaviour \cite{stange:2022,sciutti:2022,woitsch:2021}. But to be socially-aware is to be self-understanding. These robots will require using cognitive architectures that allow them not only to navigate and interact in a socially accepted way but also to be able to \cite{chakraborti:2021} (i) recognize, understand and predict the behaviour of people around the robot, to keep a model of their behaviours as well as a semantic representation of the world that allows remembering and planning using social cues; (ii) anticipate not only perceptions but also the consequences of ongoing and possible actions; and (iii) learn to adapt to different situations, contexts and users, updating their model of others by themselves. Moreover, as a relevant factor for influencing human trust, it is also important that the robot can explain the reason behind its behaviour to the human companion \cite{setchi:2020}, either at their request or on their own initiative because they feel that such explanations are necessary. Causal discovery and reasoning, as one of the current trends in artificial intelligence, is exposed as a potential tool in the management of previously mentioned requirements. From a complementary point of view, as social robots augment their autonomy, enabling them to operate in increasingly open-ended environments, enormous possibilities open for improvements in the human economy and well-being. However, it also poses strong risks that are difficult to assess and control by humans. The recent wave of artificial intelligence based on data-intensive machine learning exacerbates these risks due to (1) severe limitations in the handling of novel situations, and (2) a lack of transparency and explainability \cite{gjaerum:2023}. The efficiency, robustness, safety, adaptability and trust of AI-driven robots are at stake. Suitable technology for the construction of dependable, trustable, autonomous robots is in strong need for both (a) the core operation of the robot, and (b) the engineering process used to build it. Some of these capabilities, resilience, and trust problems of autonomous robots can be traced to robots suffering a lack of understanding of what is going on.

This paper reviews the efforts addressed for allowing a robot to create explanations about its own experience and, in particular, to explain plan updates or perceptual anomalies causally and to integrate these explanations into its body of knowledge and use them later to improve its behaviour. We can set the departure point in the problem of the robot explaining its actions when they are the result of an internal planning process. Those explanations are usually oriented to provide a humanly understandable representation for debugging purposes or to determine liability, or eligibility in autonomous operations \cite{rodriguezlera:2022,lindner:2022}. But here we will also review the different problems of providing explanations that can be associated with real communication with the person \cite{chakraborti:2017}. This is a very general approach that can be related to scientific and human insight, sometimes stated as the “coordination of claims (i.e., models) and evidence” \cite{sandoval:2015}. Always within the context of the specific goals being pursued and the value that they convey to humans, self-understanding and self-expansion refers to how human-robot communication loops operate \cite{kontogiorgos:2023} and how they can be promoted within a cognitive architecture, making them a structural part of it. Following J. Wyatt, “A system with self-understanding is any system with explicit representations that captures some kind(s) of incompleteness in some kind(s) of knowledge that it possesses. A self-extending system is a system that is able to fill in this incomplete knowledge.” \cite{wyatt:2010}. 

The paper is organised as follows: Section \ref{sec2} analyses research efforts carried out in recent years that address the problem of generating explanations to questions that a person can generate for the robot without taking into account or considering constraints. This study considers the inclusion of cause-effect relationships as the common thread in the attempt to make autonomous robots capable of providing explanations and thereby inspire more confidence in the people with whom they share their environment and, sometimes, tasks. The ability of robots to plan how to do a task is not the focus of this survey, but it is closely related to the existence of an internal causal model. The generation of explanations in a scenario where the questions to be answered do not impose conditions is considered first. The robot seeks to explain to the person the reason for the actions it is executing or is going to execute, or the reasons behind a perceptual anomaly. Secondly, we study the case in which the robot needs to give constrained explanations. In this case, the robot needs to revisit the plan and, for instance, determine what happens if instead of one action, another action would have been executed. This involves a process of internalization, which presupposes the existence of a representation and tools that make it possible. Lastly, the issue of self-understanding of perceptual anomalies is addressed. The problem involves reproducing a situation in order to repeat the perceptual experience until, in this emulated environment (imagining), the causes behind the detected anomaly can be determined. In Section \ref{sec3}, the relevant issues are discussed and future lines of research are unfolded. Finally, conclusions are drawn in Section \ref{sec4}.

\section{Climbing the ladder of causation} \label{sec2}

In the field of robotics, Deep Learning techniques and, specifically, Deep Reinforcement Learning (DRL) \cite{gjaerum:2023}, have been a true revolution, allowing robots to be endowed with both a great perceptive capacity and an advanced ability to learn through the exploration of the environment, with a strategy that the robot itself improves using trial and error techniques \cite{remman:2021,haarnoja:2018}. However, these techniques rely on models whose complexity prevents people from really understanding how decisions are being made or predictions carried out. To paraphrase Aristotle, we could say that a robot has no knowledge of a thing until it has 'grasped its why, that is to say, its cause' \cite{falcon:2023} Thus, it can be argued that this lack of ability to offer comprehensive explanations prevents the general acceptance of robotic systems in critical tasks \cite{edmonds:2019}. In their surveys on explainable autonomous robots, \citet{sakai:2022} and \citet{chakraborti:2021} emphasize that, for a robot to be able to have real communication with a person, the robot must not only superficially understand the person's instructions or questions, but technologies must be developed that allow the person and robot to understand each other's internal state. As far as the robot is concerned, they state that to do this, the robot must be endowed with the ability to estimate what the person is thinking and to present its ideas in a format that is easy for the person to interpret \cite{zakershahrak:2019}. As discussed above, if the robot can know the person's internal state, it will be able to anticipate the person's needs or adapt its behaviour appropriately. If it can make the person understand what it intends to do through explanations, the robot may truly gain the trust of the people. A robot endowed with these mechanisms is referred to as an explainable autonomous robot (XAR) \cite{sakai:2022,stange:2022}. The concept of XAR is closely linked to that of explainable Artificial Intelligence (XAI), but there is an important difference. XAI methods seek to improve knowledge and control, as well as to provide a framework to justify decisions and predictions \cite{gjaerum:2023,adadi:2018}. However, if this framework focuses more on data (data-driven explainability \cite{sakai:2022}), in XAR, the problem is typically to explain to humans the actions that the robot itself is autonomously performing in an environment shared with them.  In this case, and as \citet{sakai:2022} rightly points out, the problem is one of goal-driven explainability.

\begin{figure}[t]
\centering
  \includegraphics[width=0.4\textwidth]{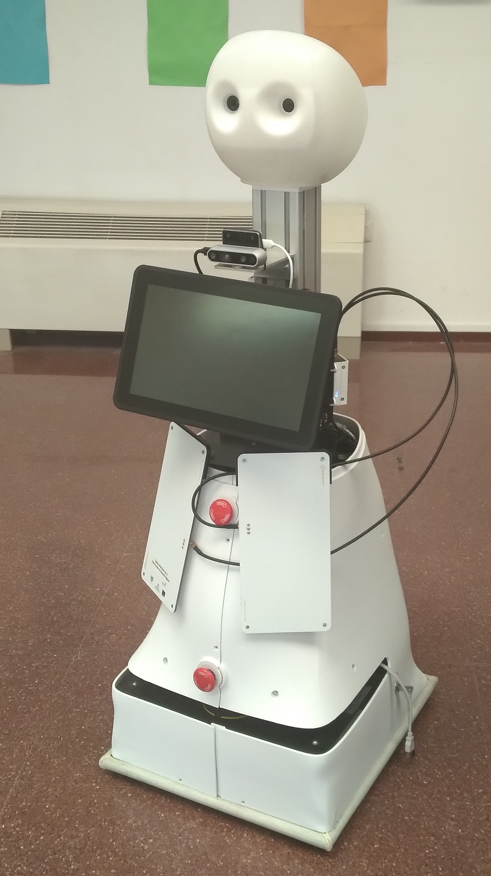}
\caption{A CLARA robot, equipped with a D435i camera from Intel, a pair of RFID antennas and a face recognition devoted hardware (Intel F455). The robot uses a SCITOS G3 as the base platform \cite{romerogarces:2022}}
\label{fig:CLARARFID}
\end{figure}

Although they are distinct features of a system, there is an intimate relationship between causality and explainability \cite{setchi:2020}. In his pioneering work, Lewis defines the act of explaining an event as 'providing information about its causal history' \cite{lewis:1986}. For Tania Lombrozo, an explanation is about filling in the gap that allows the other to understand an event \cite{lombrozo:2006}. In this sense, it is a cognitive process, in which answers are given to certain questions, usually of the "why" type and normally subject to certain constraints, since the person receiving the explanation does not want the occurrence of the event per se to be explained, but rather why it occurs in that case and not in other possible (counterfactual) cases \cite{matarese:2021}. In their causation ladder, J. Pearl and D. Mackenzie set out three steps, which define the three different levels of cognitive ability that a causal learner must possess: seeing, doing, and imagining \cite{pearl:2018}. The generation of explanations, at each of these levels, will force the robot to possess an increasingly higher cognitive level. The first involves the ability to detect regularities, and irregularities, in the environment. The second involves the ability to predict the effects of a deliberately executed action on the environment. The third allows the agent not only to know the effects caused by an action but also to know why they occur and what to do if the result differs from the expected one. The continued emergence of better sensors, the increased computing power embedded in robots, and the design of new classifiers, many based on Deep Learning techniques, have meant that, as J. Pearl and D. Mackenzie suggest, robots can now feel comfortably placed on the first step of the causation ladder. Using Deep Learning for recognising the terrain type, the hexapod robot described in \cite{chen:2024} is able to self-adapt in real-time to changes in the traversed terrain. In \cite{zhang:2022c}, gesture recognition is addressed using a double-size convolution kernel to extract feature information from images, and a two-stream convolutional neural network (two-stream CNN) to recognize the collected gesture instruction data. Figure \ref{fig:CLARARFID} shows a snapshot of the CLARA robot. This robot can detect and recognise people, talk to them, can move safely through the environment, detecting obstacles and avoiding them. The question then becomes, are they able to determine or predict the effects they cause on the environment when they decide to take a deliberate action? And if a positive answer can be given to this question, are there proposals that allow a robot to imagine what might have happened if the action taken had been something other than the one executed?

\begin{figure}[t]
\centering
  \includegraphics[width=0.7\textwidth]{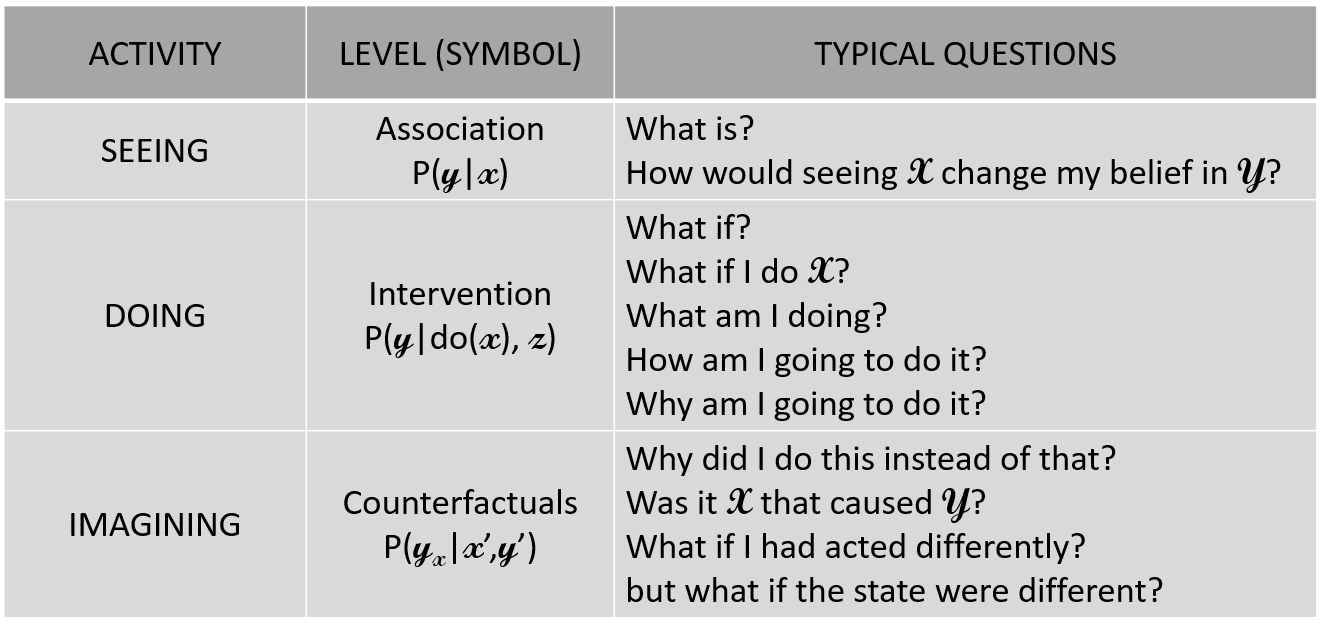}
\caption{The three rungs in the Causality ladder (adapted from \cite{pearl:2018b})}
\label{fig:tablaPearl}
\end{figure}

In addition to the activity already mentioned, the three rungs of this causality ladder can also be named according to how the agent is able to organize and manage knowledge. Figure \ref{fig:tablaPearl} is an adaptation of the one proposed by \citet{pearl:2018b}. Significantly, the figure includes, next to activity and strategies for organising or managing knowledge, the questions that can typically be answered at each step. The first of the steps is called Association, as it only invokes purely statistical relationships. The questions that the robot might ask do not require causal information. Given readings from a laser, the robot can try to estimate its pose, or the probability of colliding with an obstacle, if it maintains a speed of movement. The success of probabilistic robotics rests, in part, on conditional probability: The probability of a random variable $X$ to take a value $x$, conditioned on a second random variable $Y$ with value $y$ \cite{thrun:2005}. The second step, Intervention, involves not only perceiving what something is, but also acting in the environment. Thus, in this step, we have expressions such as the probability of the event $Y=y$ given that we act and set the value of $X$ to $x$ and subsequently perceive the event $Z$ equal to $z$. This ability of the robot to intervene is the starting point for the explanation in the work of \citet{sakai:2022}. Thus, they establish three types of questions associated with the problem of explanation. The question of "What are you going to do?" appears as the first and basic one, since it can be considered as the starting point for the other two: "How are you going to do it?" and "Why are you going to do it?". The first of these can be answered using causal reasoning and causal chains \cite{lagnado:2008,hilton:2010}. As discussed above, the response seeks to give liability or eligibility to an autonomous activity \cite{lindner:2022}. The "Why?" question is, according to \citet{sakai:2022}, the most complex or difficult to answer, and its answer must also be adapted to the intentions of the person asking it. 

In any case, all formulations at the Intervention step respond to explain the decisions that the robot has made based on knowledge of the environment that, temporally, responds to the present time. The robot is seeking to achieve a goal ("what?"), it has a plan to achieve it ("how?"), and the actions that make up this plan respond to current knowledge about the context ("why?"). The justifications that the person may ask the robot for may be due to the person's lack of knowledge of how the planner that drives the robot's behaviour works internally or to changes in the plan associated, for example, with perceptual anomalies. However, the person's questions may force the robot to abstract or explain past actions. If the robot, when it is developing the established plan, modifies it to adapt to changes in the context, the question might now be along the lines of "Why did you do this instead of that?" or, more generally, the person might ask "but what if the state were different?" \cite{gjaerum:2023}. Drawing a parallel with the two previous steps, the expressions used at this level respond to the probability that event $Y$ is equal to $y$ would be observed had $X$ been $x$, given that we actually observed $X$ to be $x$', and $Y$ to be $y$' \cite{pearl:2018b}. Significantly, we are mainly focusing the problem on what some authors call Explainable Planning. But it is not all about actions or tasks. Explanations can also have their origin in an unexpected failure or perception, which has arisen during the execution of the plan and that has changed this \cite{khana:2023,das:2021}. In many of these situations, the robot is forced to revisit the plan to determine what might have happened. However, when the robot understands that an unexpected change of state may have been due to a geometrically rooted perceptual anomaly, it may have to resort to an internalisation of the situation to explore the problem, for which the use of simulators may be required to imagine what might have happened \cite{beetz:2018,Bustos2023}. 

Next Sections provide further details about different methods of knowledge representation in autonomous robots and how they contribute to explainability, and also implemented approaches for providing an answer to the questions in the Intervention and Counterfactuals steps of the causality ladder. As will be discussed in Section \ref{planning}, in its simplest version, to answer the Intervention-related questions, the robot must be able to draw up a plan to achieve a goal, concatenating actions that are triggered by the fulfilment of preconditions. The robot could use this information to answer the "how?" and "why?" questions. To answer the Counterfactuals-related questions, the robot should understand what should be changed in the input instance to change the result obtained. These are post-hoc local explanations \cite{guidotti:2022}.

\subsection{Knowledge representations and explainability} \label{KR}


Table \ref{tab:KR} summarises some typical examples of knowledge representations used by different authors as a basis for enabling an autonomous robot to generate explanations of its behaviour. The table provides reference to these works, the knowledge representation scheme used, the position we could assign to the proposal in the causality ladder (Figure \ref{fig:tablaPearl}), and the applications in which these proposals are used.

\begin{table}
\centering
\caption{Knowledge representation in autonomous robotics and their relationships with explainability}
\begin{tabularx}{\textwidth}{|>{\raggedright\arraybackslash}p{0.19\textwidth}|>{\raggedright\arraybackslash}X|>{\raggedright\arraybackslash}p{0.16\textwidth}|>{\raggedright\arraybackslash}X|}
\hline
Approach & Knowledge representation & Step (Figure \ref{fig:tablaPearl}) & Application \\
\hline
\citet{gao:2020} & And-Or graph & Intervention & Human-robot collaboration tasks \\   
\hline
\citet{han:2021} & Behaviour Trees & Intervention & Explainable planning \\       
\hline
\citet{lindner:2022} & PDDL domain and E/D (causal) links & Intervention & Explainable planning \\       
\hline
\citet{krarup:2021} & extended PDDL2.1 & Counterfactuals & Explainable planning \\        
\hline
\citet{diehl:2022} & Bayesian networks & Intervention & Failure explanation \\        
\hline
\citet{stange:2022} & A collection of memories and models & Intervention & Human-robot interaction \\       
\hline
\citet{olivares:2021} & Ontology for Collaborative Robotics and Adaptation & Intervention & Human-robot collaboration tasks \\        
\hline
\citet{hanheide:2017} & 3-layers hierarchical organization & Counterfactuals & Task-planning and execution in open and uncertain worlds. Failure explanation \\
\hline
\citet{gjaerum:2023} & Linear Model Trees & Counterfactuals & Getting counterfactuals explanations in real-time for black-box models \\ 
\hline
\end{tabularx}
\label{tab:KR}
\end{table}

\begin{figure}[t]
\centering
  \includegraphics[width=0.6\textwidth]{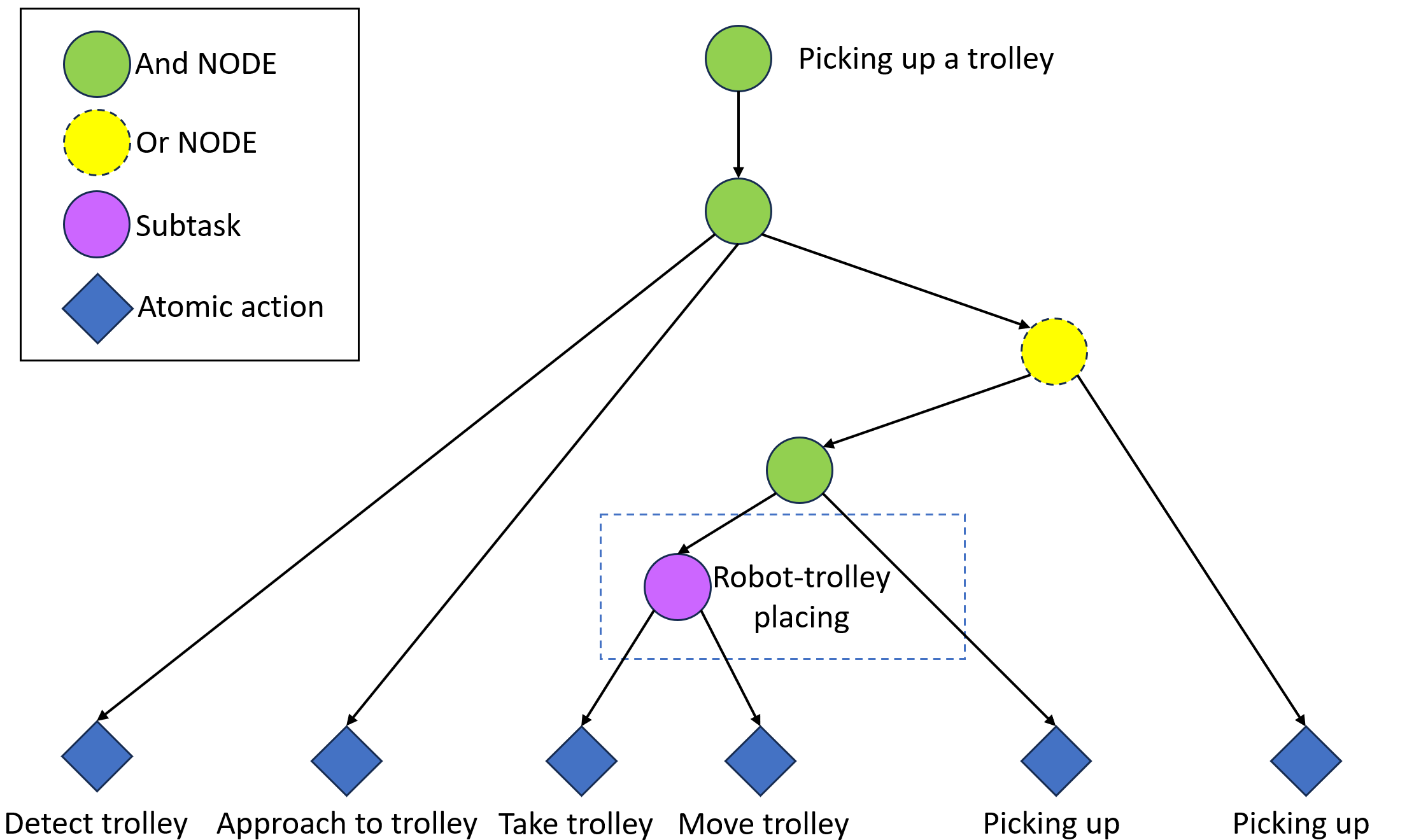}
\caption{The hierarchical mind model for a collaboration task represented by a spatial-temporal-causal And-Or graph \cite{gao:2020}. The task "Picking up a trolley" implies that the robot must detect and approach to the trolley. Once the robot is correctly positioned close to the trolley, it can either ask the person to manually place the trolley on its pallet truck (because, for example, space constraints prevent the robot from doing so) or it can move and pick it up itself. }
\label{fig:AndOrGraph}
\end{figure}

In general, it can be said that much of the work in this field has focused on the aforementioned Explainable Planning. Therefore, many of the representations seek only to describe the course of action, and use objects or locations as attributes of the tasks to be performed. This is, for example, the case of \citet{gao:2020}'s proposal, in which so-called And-Or graphs are used. A simple example is shown in Figure \ref{fig:AndOrGraph}. The goal in this work is to represent a task that is carried out collaboratively, and in which the robot generates explanations about the progress of the plan and the next task to be performed. The model and internal parameters are adjusted by hand, although the authors leave open the option of learning from annotated user data. The decision-making space can be also encoded as a graph. \citet{zhang:2021} propose to represent the model as a sparse multi-step graph where nodes are representative landmarks. These landmarks are scattered (in terms of reachability) across the goal space, and learnt via clustering a latent space built using an auto-encoder with reachability constraints. Similarly, in the proposal by \citet{hu:2023}, the robot learns a graph-structured world model consisting of sparse, multi-step transitions. Policy nodes based on reachability estimates in the global space are generated. Thus, planning in a continuous state space is simplified to a graph search problem. 

Other schemes, popularly used to represent and decompose a task into lower-level objectives, have also been used in frameworks that consider explainability. This is the case of Behaviour Trees (BTs). In the proposal by \citet{han:2021}, high-level robot explanations are encoded using BTs, framing them into a set of semantic sets (the goal, subgoals, steps, actions). They provided contributed algorithms to generate robot explanations, specifically giving answers to questions seeking causal information. However, as discussed in more detail in Section \ref{planning}, the advantages offered by Automatic Planning make it very common in the context of explainable planning to work with Planning Domain Definition Language (PDDL). Criticisms of the use of PDDL for symbolic planning, due to difficulties in scaling in a real environment \cite{capitanelli:2023}, do not correspond to the results provided by different works \cite{pulido:2017,li:2021b}. The use of PDDL allows the plan to be synthesised action by action rather than end-to-end as is the case when using deep learning algorithms, which makes it easier to modify it at runtime when the environment changes. Moreover, the latter allows each action to be available for execution as soon as it is generated, which in turn allows concurrent planning and execution. PDDL is used, for example, for the proposals of \citet{krarup:2021} or \citet{lindner:2022}, which are cited in Table \ref{tab:KR}, but also for \citet{sreedharan:2021} or \citet{stulp:2020}. In general, the coding of preconditions and postconditions used in PDDL constitutes a causal model that can be quite complex and that can allow work on reconciliation between the robot and person models \cite{sreedharan:2021}, or on post-hoc explanation \cite{krarup:2021} (counterfactuals, i.e. , an explanation that describes that the output of the model will be $y'$ instead of $y$, thereby changing, without this change occurring in the real world, the behaviour or inputs $x$ to $x'$ \cite{verma:2021}

To address the problem of explaining the reasons behind functional failure, \citet{diehl:2022} propose to learn a Bayesian causal network from simulations. The model is then successfully transferred to real environments. The generation of explanations is based on setting the failure state in contrast with the closest state that would have allowed for successful execution. The representation is scalable to manage other problems or platforms. 

The cognitive and social sciences have long studied how people explain their own behaviour to others. This research has served as the basis for proposing explanatory frameworks in robotics, which thus emphasise the social-interactive and human behavioural aspects of explanation. \citet{stange:2022} propose a framework where several types of memories coexist (working memory, episodic memory). A state machine is encoded as Interaction model, allowing to recognize and keep track of the human-robot interaction context. The aim is to give the robot the ability to automatically generate verbal explanations of its own behaviour, mapping its internal courses of action for decision-making. 

When trying to capture the whole reality of a scenario, without limiting it to a specific application or restricting it only to task management, the knowledge representation must be able to combine different types of tasks, robots, objects, environments... A viable strategy to consider all this information is the use of knowledge-enabled approaches based on ontology \cite{manzoor:2021}. These ontology-based approaches can be adapted to deal with evolving and changing scenarios and different applications, facilitated by the fact that pieces of knowledge, which are in many cases independent of each other, can be reused. There are different semantic knowledge representations in robotics: RoboEarth \cite{waibel:2011}, KnowRob \cite{beetz:2018}, RoboCSE \cite{daruna:2019} The KnowRob architecture allows the robot to emulate a situation, so that it can draw up a plan of action that it can then execute in reality.  Although it does not specifically have a module for generating explanations, its internal functioning based on questions and answers, which the robot asks itself to a certain extent, is very interesting. The ARE-OCRA \cite{olivares:2021} is a framework for generating explanations about target queries. The knowledge base is built using an Ontology for Collaborative Robotics and Adaptation (OCRA).

Some representations are specifically designed to manage knowledge in the application framework addressed by the proposal. For instance, the IID schema proposed by \citet{hanheide:2017} stores knowledge about the state and how an action modifies it. By state is meant specific locations and categories of objects, rooms, and places in the building, as well as the whereabouts of particular people and what they know. It organises this knowledge in three layers: instance, default, and diagnostic. The instance layer contains state information. The default layer manages general knowledge about categories (e.g. objects typically found in office rooms) and about the effects of an action on the instance layer. The diagnostic layer contains the action knowledge being responsible for creating new general knowledge and  augmenting the action models in the default layer with possible causes of failure. This knowledge representation is the core of a cognitive architecture, where the flow of information is controlled by employing a distributed blackboard. Task-related agents (dialogue, mapping) are distributed within this architecture in a lower layer. A second layer is included for generating the symbolic items in the knowledge representation. These symbols are used for decision-making in the higher layer of the architecture. The existence of a central blackboard as a working memory extended with an internal simulator is the core of the proposal in \citet{Bustos2023}. The aim is to endow robots with the ability to navigate and follow those humans being assisted by them, while verifying the coherence between the perceived reality and the simulated one. Inconsistencies are detected triggering the search for causal explanations and the acquisition of new concepts.

\begin{figure}[t]
\centering
  \includegraphics[width=0.8\textwidth]{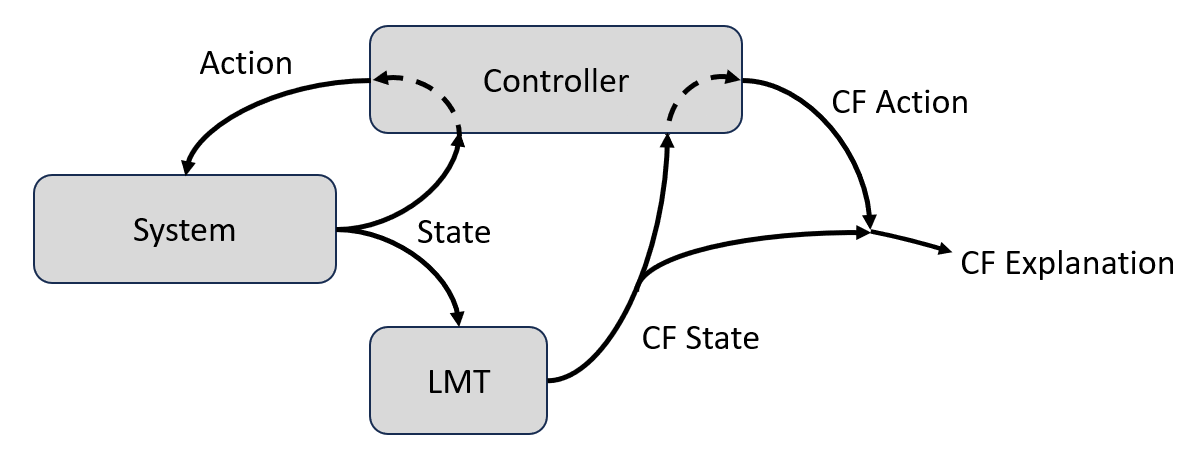}
\caption{Schematic view of the counterfactual explanation generation in the \citet{gjaerum:2023}'s proposal (see text for details)}
\label{fig:gjaerum}
\end{figure}

There are far fewer proposals to address the problem of reasoning, or providing explanations, for a plan that has not actually been implemented. \citet{gjaerum:2023} propose to use linear model trees (LMTs) for this purpose. The idea is to get real-time counterfactual explanations for a black-box model using the LMT as a surrogate model. Figure \ref{fig:gjaerum} shows how this scheme would work. Basically, the controller captures the state of the system and returns the appropriate action to execute. The LMT captures that same system state and computes what the counterfactual state would be, which is provided to the controller. With this input, the controller provides a counterfactual action. The combination of state and counterfactual action provides the explanation. Given the low level of the response (the result of a state-action correlation), this system applies to equally low-level controllers. 

\subsection{Explaining a plan: how and why are I going to do it?} \label{planning}

Current robotics solutions need to deal with a dichotomic situation: on the one hand, it is expected that they can operate in open-ended environments, but, on the other hand, the complex tasks that they must fulfil are typically specified using static Finite-State Machines or Behaviour Trees. Thus, they are “mostly” blind concerning the future; they usually ignore the impact of the selected state on the next reasoning processes. Moreover, in their standard implementations, they are free-form and static solutions to task planning \cite{han:2021}. As discussed in the previous section, automated solutions do exist to allow for performance planning, rather than mere imitation. Can a robot predict the effects of a deliberatively executed action? Automated Planning \cite{ghallab:2004} is particularly suitable in robotic systems because it employs some form of forward projection (reasoning in depth about goals, preconditions, resources, and timing constraints) and provides the robot reasonable responses in states unforeseen by the designer \cite{cashmore:2015,rajan:2012}. The comparison of the behaviour of a planner in a scenario where the robot shares its environment, and perhaps task, with people, will be based on objective measures (such as efficiency in task execution, accuracy or time spent), and subjective measures, which can be captured with Likert-scale questionnaires \cite{narayanan:2015} These questionnaires will assess determinants such as situational awareness, complacency, likability, trust, or mental workload. The aim of providing explanations should be to improve these metrics. The topic is addressed in detail in \citet{krarup:2021}'s work.

Typically stated as a Markov Decision Process, a decision-making space $\Pi$ can be expressed using the following transition function \cite{chakraborti:2021,sakai:2022}
\begin{equation}
\delta_{\Pi} : A \times S \rightarrow S \times \Re,
\end{equation}
where $\Pi$ is the decision-making space, $A$ and $S$ are the actions available to the robot and the set of possible states it can be in, respectively, and $\Re$ represents the cost (expressed as a real number) required for the transition. Using this terminology, the plan $\pi$ is a sequence of actions
\begin{equation}
\pi = \{ a_1, a_2, ... a_i\},~~a_i \in A,
\end{equation}
which is generated using an algorithm $\mathcal{A}$ and under a certain constraint $\tau$, i.e. 
\begin{equation}
\mathcal{A} : \Pi \times \tau \rightarrow \pi,
\end{equation}
Briefly, the plan transforms the current state $X \in S$ to its goal $G \in S$, i.e. \begin{equation}
\delta_{\Pi} (\tau, X) = \langle G, \sum_{a_i \in \pi} c_i\rangle,
\end{equation} 
where $\sum_{a_i \in \pi} c_i$ denotes the plan cost $c(\pi)$ \cite{das:2021}. The policy can be expressed as
\begin{equation}
\pi : s \rightarrow a, ~~~~ \forall a \in A, ~~~ \forall s \in S.
\end{equation}

\begin{figure}[t]
\centering
  \includegraphics[width=0.6\textwidth]{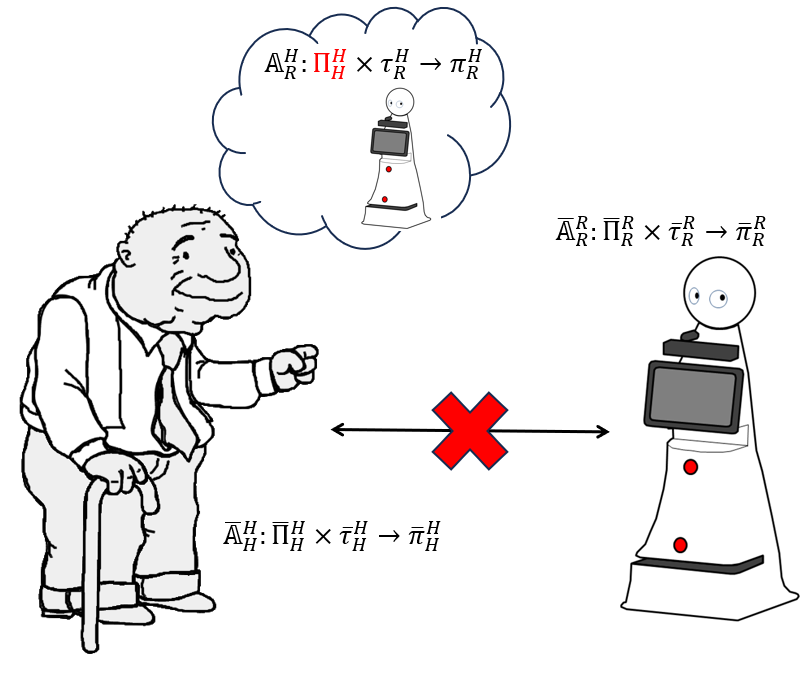}
\caption{Person and robot are internal plans $\overline{\pi}^R_R$ and $\overline{\pi}^H_H$, respectively, generated with respect to their internal models. The person has no information about the robot's model. Or the robot is not explaining (or correctly explaining) its plan to the person. In any case, the problem is that the robot's plan is interpreted by the person (${\pi}_R^H$) with respect to his model ${\Pi}^H_H$  \cite{chakraborti:2017}}
\label{fig:badXAR}
\end{figure}

Explanations typically justifies to the person that the solution $\pi$ satisfies $\tau$ for a given $\Pi$. Correctly, \citet{das:2021} add a new explanation type: that which allows the person to understand that a wrong state has been reached unexpectedly, which has meant stopping the execution of the plan $\pi$. In any case, the most basic explanation of a plan consists of the planner explaining the decisions it has made based on its internal model (i.e. current state, actions and goals) \cite{lindner:2022,chakraborti:2017}. In the Cambridge Dictionary, this form of expressing oneself is called soliloquy, and is defined as "a speech in a play that the character speaks to himself or herself or the people watching rather than to the other characters". Obviously, these explanations will only be valid if the person receiving them is intimately familiar with the internal model of the robot, with how the robot organizes knowledge and how the planner decomposes the execution of a plan. Otherwise, when the human interlocutor is a non-expert user, possessing a domain and task model that differs significantly from those internalized by the robot, the validity of these explanations may be far from being of any use. Figure \ref{fig:badXAR} schematizes this situation. The terminology used follows that proposed by \citet{sakai:2022}. Both the robot and the person are carrying out their plans. Assuming that the decision-making process can be applied to people and robots, we can express these internally generated policies as
\begin{equation}
\begin{split}
\overline{\mathcal{A}}^H_H : \overline{\Pi}^H_H \times \overline{\tau}^H_H \rightarrow \overline{\pi}^H_H \\
\overline{\mathcal{A}}^R_R : \overline{\Pi}^R_R \times \overline{\tau}^R_R \rightarrow \overline{\pi}^R_R
\end{split}
\end{equation}
where $\overline{\mathcal{A}}^H_H$ refers to the person's cognitive model of its own needs, behaviours, goals..., and $\overline{\mathcal{A}}^R_R$ to the robot's computational model of its own behaviours, goals.... Being internally generated, they are not directly observable by the others. The interpretable policies can be expressed as
\begin{equation}
\begin{split}
{\mathcal{A}}^H_H : {\Pi}^H_H \times {\tau}^H_H \rightarrow {\pi}^H_H \\
{\mathcal{A}}^R_R : {\Pi}^R_R \times {\tau}^R_R \rightarrow {\pi}^R_R
\end{split}
\end{equation}
Internally, the person interacts with a model of the robot, $\mathcal{A}^H_R$. This interaction process can be expressed as
\begin{equation}
{\mathcal{A}}^H_H : {\Pi}^H_H \times {\tau}^H_H \rightarrow {\pi}^H_H \rightleftharpoons
{\mathcal{A}}^H_R : {\Pi}^H_R \times {\tau}^H_R \rightarrow {\pi}^H_R
\end{equation}
But for this to be possible, the plan generated by the robot concerning its model needs to be interpreted by the person (${\Pi}^H_R$). However, as it is shown in Figure \ref{fig:badXAR}, if the robot does not provide or provides insufficient explanations (or its behaviour does not make sense with respect to the person's model), this will be interpreted by the person with respect to her own model ($\Pi^H_H$). Although we consider in the figure that the human understands the robot's constraints (they are an estimate, $\tau^H_R$, of $\tau^R_R$), further explanations are needed. 

\citet{sakai:2022} identifies four requirements for allowing a robot to generate an explanation. Briefly, the internal model of the robot must be interpretable by the person, thus avoiding that the robot behaves strangely for the person; and the person and the robot must have adequate versions of the elements that define the other's plan (model, constraints, plan generation algorithm). Finally, the explanations provided by the robot must be understandable to the person, avoiding verbalization or visualization problems. Figure \ref{fig:badXAR} shows the problem that arises when one, or more, of these requirements are not met. The result shown in the figure may be the result of a robot whose behaviour is not reasonable or understandable to the person (it may be caused by the internal model used by the robot). But the problem could also be that the internal models used by the robot and person, $\overline{\Pi}_R^R$ and $\overline{\Pi}_H^H$ respectively, are very different (actions, states...). This is more common when the physical capabilities of the robot are very different from those of the person and, as aforementioned, when the person is a non-expert \cite{das:2021,raman:2013}. To deal with this situation, \citet{raman:2013} propose to use Linear Temporal Logic. But in this proposal, explanations focus on which actions are infeasible for the robot, but not on why they are not. And therefore, these explanations are often insufficient. \citet{das:2021} propose a new type of explanation, which describes the cause of unexpected failures during the execution of the plan. They show that explanations capturing the context of a failure and also the history of past actions, are the most effective for failure and solution
identification among non-experts. The four requirements for achieving an explainable robotic agent are summarized in Table \ref{table1}. 

\begin{table}[H]
\centering
\caption{Issues for achieving an explainable robot \cite{sakai:2022}}
\begin{tabularx}{\textwidth}{|>{\raggedright\arraybackslash}p{0.30\textwidth}|>{\raggedright\arraybackslash}X|}
\hline
Requirement & Solutions \\
\hline
The sequential decision-making $\Pi^R_R$ must be interpretable by the person. That is, given $\mathcal{A}^R_R$ and ${\tau}^R_R$  
\begin{equation}
{\mathcal{A}}^R_R : {\Pi}^R_R \times {\tau}^R_R \nrightarrow \emptyset
\end{equation}
When $\pi^R_R$ is derived using these and ${\Pi}^R_R$, without encountering an error, the decision-making space is defined as interpretable.
& 

$\Pi_R^R$ must be composed of states and actions natural to people, with transitions between states that are consistent with the real world. One solution is to manually encode the model using a collection of rules, but it is obviously complex that this set will be complete \cite{sakai:2022}. The solution can be to build a causal graph \cite{madumal:2019,diehl:2022}. For instance, \citet{stocking:2022} analyse how to distinguish task-relevant and -irrelevant variables exploring causality. \citet{brawer:2020} propose an approach for a robot to learn an explicit model of cause-and-effect by constructing a structural causal model. \citet{diehl:2022} describe an approach for learning a causal Bayesian network from simulated task executions. The obtained model was used for transferring the acquired knowledge from simulation to several real robots with different embodiments.\\
\hline
The robot must possess the tools ($\Pi^R_H$, $\mathcal{A}^R_H$, and $\tau^R_H$) that allow it to infer the person's behaviour. 

\begin{equation}
{\mathcal{A}}^R_H : {\Pi}^R_H \times {\tau}^R_H \rightarrow \pi^R_H
\end{equation}
where $\pi^R_H$ is a close estimate of $\pi^H_H$.

& 

The Bayesian framework has been extensively used to predict human reasoning in goal-oriented tasks \cite{jaraEttinger:2016,huang:2019}. \citet{gao:2020} propose to understand complex human activities using an action parsing algorithm based on a graph-based task representation. This graph allows the robot to infer human mental states in complex environments.\\
\hline
The robot must be able to generate explanations that allow the person the comprehension of the robot's plan $\pi_R^R$. This explanation should allow the person to have an estimate, as close as possible, to the plan executed by the robot, but not necessarily to understand the robot's policy. 

& 

Explanations must achieve a real interpretable decision-making space. In Figure \ref{fig:badXAR} the situation has been exemplified in a very drastic way: the model that the person uses to interpret the robot's plan is not influenced by the robot itself. Assuming that there is some influence (and that the model of the robot's behaviour is estimated in some way by the person, $\Pi_R^H$), and denoting by $\epsilon$ the explanation provided by the robot, the objective is that $\Pi_R^H+\epsilon$ allows the person to internalize a model $\widehat{\Pi}_R^H$ such that 
\begin{equation} \label{MRP}
\mathcal{A}_R^H : \widehat{\Pi}_R^H \times \tau_R^H \rightarrow \pi_R^R. 
\end{equation}
Different approaches have been proposed to generate explanations in the field of human-robot interaction. Typically, these works seek to provide explanations in a particular order or to interweave them with the execution of the plan. Since the person does not always want the same level of detail in the explanation, \citet{zakershahrak:2020} propose a hierarchical model of explanations and details that the robot requires to provide explanations as a target. \citet{han:2021} propose to encode high-level robot explanations using Behaviour Trees (BTs), specifically giving answers to questions seeking causal information. In the proposal by \citet{gao:2020}, the internal model of the person is inferred and compared with the robot's model. Whenever a disparity between these two models is detected, an explanation is generated to encourage a correction of the person's internal state. \\
\hline
Explanations are not correctly encoded/conveyed to the person & We must note that, solving Equation (\ref{MRP}), is only the first step for providing an explanation. That is, after determining the explanation $\epsilon$, the robot needs to encode it into a form that can be conveyed to people. This problem can be solved using verbalization \cite{bradley:2017} or visualization \cite{elhafi:2020} schemes. But, in general, the goal should be to personalise the communication channel according to the preferences of the end-users. In \cite{romerogarces:2020}, the human-robot interaction employs several informative channels and this personalisation is achieved by using an adaptation framework. It is also relevant how the information is presented to the human. Most work on explanation generation assumes that the person understands an explanation, regardless
of how much information it includes. As \citet{zakershahrak:2019} pointed out, complex
explanations should be provided in an online fashion, broken each explanation into multiple parts, which are
then provided to the person separately and intertwined with the execution of the plan.
\\
\hline
\end{tabularx}
\label{table1}
\end{table}

The first requirement implies that the robot must internalise a model that is interpretable by the person. The goal is to provide good explanations for the person, not for the robot itself. However, several approaches focus on providing good explanations from the perspective of the robot \cite{hanheide:2017}. To alleviate this problem, some work actually assumes that the models for the person and the robot can be the same. In this case, an explanation from the robot's perspective would not only be correct, but also perfectly understandable for the human \cite{zakershahrak:2021}. As the references shown in Table \ref{table1} show, it is very common to employ hierarchical models to represent the plan (And-Or graphs, Behaviour Trees) which, in collaborative frameworks, encode both the one maintained by the robot and the one inferred by the person.
For instance, in the proposal by \citet{gao:2020}, a spatial-temporal-causal And-Or graph is employed for encoding the joint plan (Figure \ref{fig:AndOrGraph}). Significantly, this graph encodes a joint task plan, providing an unified representation of the robot's knowledge and plan as well as the inferred person's knowledge and plan. During the collaboration, the robot manages an instance of this graph to represent its internal state and another one to represent the inferred person internal state. They shared the same structure, and, at a given time $t$, the robot can compare the most possible person mental state with its internal state. If they are the same, there is no need to explain to the person. In the other case, the robot should provide the explanation. As described in Section \ref{KR}, the decision-making space can be also encoded as a graph \cite{zhang:2021,hu:2023}.

However, in a more general case, it is assumed that both models differs. In the context of Automated Planning and Robotic systems, Problem Modelling Languages such as the aforementioned PDDL \cite{mcdermott:1998,haslum:2019} constitute an attempt to standardize the knowledge representation of planning problems using a high level of abstraction. They allow us to model the physical states, the transformations between these states, the goals and the requirements of the robot tasks in a declarative form, regardless of the robotic platform or the specific programming language used in the low-level modules of the architecture. The formulation of planning as a state transition system has been at the core of domain-independent planning research and, specifically, of PDDL. Thus, although several extensions have been devised to extend the expressiveness of PDDL \cite{fox:2003}, the focus remains on handling causal relationships derived from the conditions and effects of primitive actions. Equation (\ref{MRP}) was initially proposed by \citet{chakraborti:2017} as a model reconciliation problem (MRP), i.e. the problem of bringing the human's model closer to the robot's model by making minimal changes. In their work, PDDL was used to state and analyse the MRP. Given the high-level description in PDDL of the domain, the problem and the goals to be reached, the planning systems aim to find a sequence of actions to be executed by the robot to reach those goals. In this case, the planning system provides a flow of actions that is adapted in runtime to the changes in the environment. Regarding the latter, contextual information can be considered as part of the state description of the planning system. Depending on the runtime estimation of these context-related metrics, the planning system could decide to adopt the nominal course of action. To implement this idea, the planning system should manage in a high-level of abstraction the execution of the low-level actions, which will be distributed in the task-related modules of the robotic architecture. Therefore, it can be noted that the use of Automated Planning in robot control will require the definition of at least two levels of abstraction: high-level (deliberative planning), and low-level (robot sensing and reactive behaviours). The mapping  between these two levels of abstraction is required to solve real tasks \cite{coradeschi:2003,bustos:2019, marfil:2020}. Typically such mappings are written in a specific programming language by experts who know the software, in an ad hoc manner for each particular robot platform and task planning system. In this way, including modifications requires, on the one hand,  to resort to experts who know the software architecture and, on the other hand, rewriting the source code, and recompiling the entire software architecture in the worst case. However, ideally, such mapping could be also defined in a declarative way, so that they can be easily edited (even by non-experts) and without modifying the modules of the control architecture, which should be able to process such formal description. 

\subsection{Exploring the recent past: why did you do this instead of that?}

As reviewed in the previous section, the need for an explanation typically arises when a mismatch is detected between the plan being executed by the robot and the one expected by the person. To try to find out where the difference between the plans lies, the person expresses his or her doubts with questions.  However, several studies \cite{gjaerum:2023,guidotti:2022,krarup:2021} have shown that, when people ask questions about plans, those questions are contrastive, i.e. "Why did you do this instead of that?". The difference between this question and the question "Why do you do this?" is relevant, as it involves removing ambiguity to some extent. A contrastive question aims to include the context of the question, providing a more realistic idea of what the questioner needs in the explanation. For instance, in \citet{krarup:2021}'s proposal, this situation is posed as a plan negotiation problem. Thus, in order to get the person to understand the robot's plan, an iterative process of questions and answers (explanations) arises, the ultimate goal of which is that the person is able to understand this plan $\pi_R^R$. A domain-independent approach is proposed to compile these questions into constraints is proposed. These constraints are added to the planning model, so that a solution to the new model represents the contrastive plan that is used to produce an explanation.

Contrastive questions can be considered as local or global \cite{mueller:2019}. Local questions are asked when the person wants an explanation of specific decisions made by the robot at a given time, while global questions are asked when one wants to better understand how the robot makes decisions in general. In both cases, they can be restricted to a specific model. In one case, questions are asked to find out why a decision has been made (or what could be expected if the decision had been different), while in the global case, questions seek knowledge about the model itself. In this paper we will focus on the more local framework: the person has a relatively deep knowledge of the model, and the questions are asked to find out details in the execution of the plan or to corroborate that the plan was adequate (or where it might have gone wrong). However, this is not to say that many of the methods discussed in this study are not capable of being employed in different scenarios. In the specific case of the method proposed by \citet{krarup:2021}, the idea is that the approach implemented to compile the issues into plan constraints (and thus solve the plan negotiation problem) is domain-independent. The example described in the article could easily be a different one. The same is true for the Linear Model Trees approach proposed by \citet{gjaerum:2023}, which is in fact applied in this work in two very different scenarios.

Counterfactual explanations attempt to answer hypothetical questions, which involve revisiting the recent past to analyse states or actions \cite{smith:2020}. Although they could be posed in the form: "But what would you have done if the state had been different?", the question typically takes the form of "Why did you do this instead of that?". In the typical examples used to describe what a counterfactual explanation is, the emphasis is on the importance of providing actionable explanations, i.e. explanations that would enable the recipient to change certain parameters and thereby successfully achieve the desired outcome. This is certainly the ultimate goal of explaining. However, as \citet{gjaerum:2023} rightly describes, in Robotics, it does not make much sense for an explanation to seek that you change the input parameters (the state), but rather that they are feasible, i.e. that both the counterfactual state and the counterfactual action are physically possible \cite{guidotti:2022}. 

To provide these post-hoc explanations, \citet{carreira:2021} propose a method based on classification trees with both univariate and multivariate splits in the branch nodes. \citet{karimi:2019} employ a method in which finding these explanations is expressed as a sequence satisfiability problem. Neither of these two solutions can be applied to regression problems with multiple outputs, which makes it difficult to apply them to the complex scenarios in which a real human-robot interaction process will be framed. The concept of the leaf-to-leaf counterfactual distance matrix is proposed by \citet{sokol:2019} to describe how much an instance belonging to a leaf node must be changed to belong to a different node. The framework is again classification trees. \citet{gjaerum:2023} propose to use decision trees with linear functions in the leaf nodes (linear model trees \cite{gjaerum:2023b}), showing that they can be valid for generating counterfactual explanations in relatively competitive scenarios with multiple, continuous outputs. Specifically, the challenge is solved using H-LMT (Heuristic Linear Model Trees) \cite{gjaerum:2021}.  A H-LMT does not take maximum depth as a restriction but rather the maximum number of leaf nodes allowed. This allows to grow asymmetric trees that grow deeper in the more complex parts of the feature space while leaving the parts of the tree covering simpler regions shallow. \citet{krarup:2021} set up the problem of generating an explanation as an iterative process, in which the person repeatedly asks the robot. These questions are contrastive in nature, and each of them leads to a more constrained planning problem (e.g., the question "Why did you do A instead of B?" involves drawing a plan in which B is in place of A). Solving this new planning problem allows the person to compare and contrast the alternative proposed by the person with respect to the original plan, and allows the person to gain knowledge about the robot's internal model and the robot to gain knowledge about the person's preferences. This proposal formalises the so-called explanation as exploration, based on the idea that the person interested in learning more about how a black-box model works will ask contrastive questions in an iterative way \cite{hoffman:2018}. 

\subsection{(Self)Explaining perceptual anomalies}
 
Elaborating on the previous section, a more specific need for explanation arises when the robot experiences a perceptual anomaly. An anomaly can be defined as a substantial deviation in the perceptual input from a model-based prediction. For instance, we can imagine a scenario in which a robot follows a human, carrying some object for her. At some point, the robot hits an unforeseen pothole and the object tips over. We assume that both the tilting caused by the pothole and the falling of the box can be detected by its sensors, let's say an Inertial Measurement Unit (IMU) and a camera. 

\begin{figure}[h]
    \centering
    \includegraphics[width=0.7\linewidth]{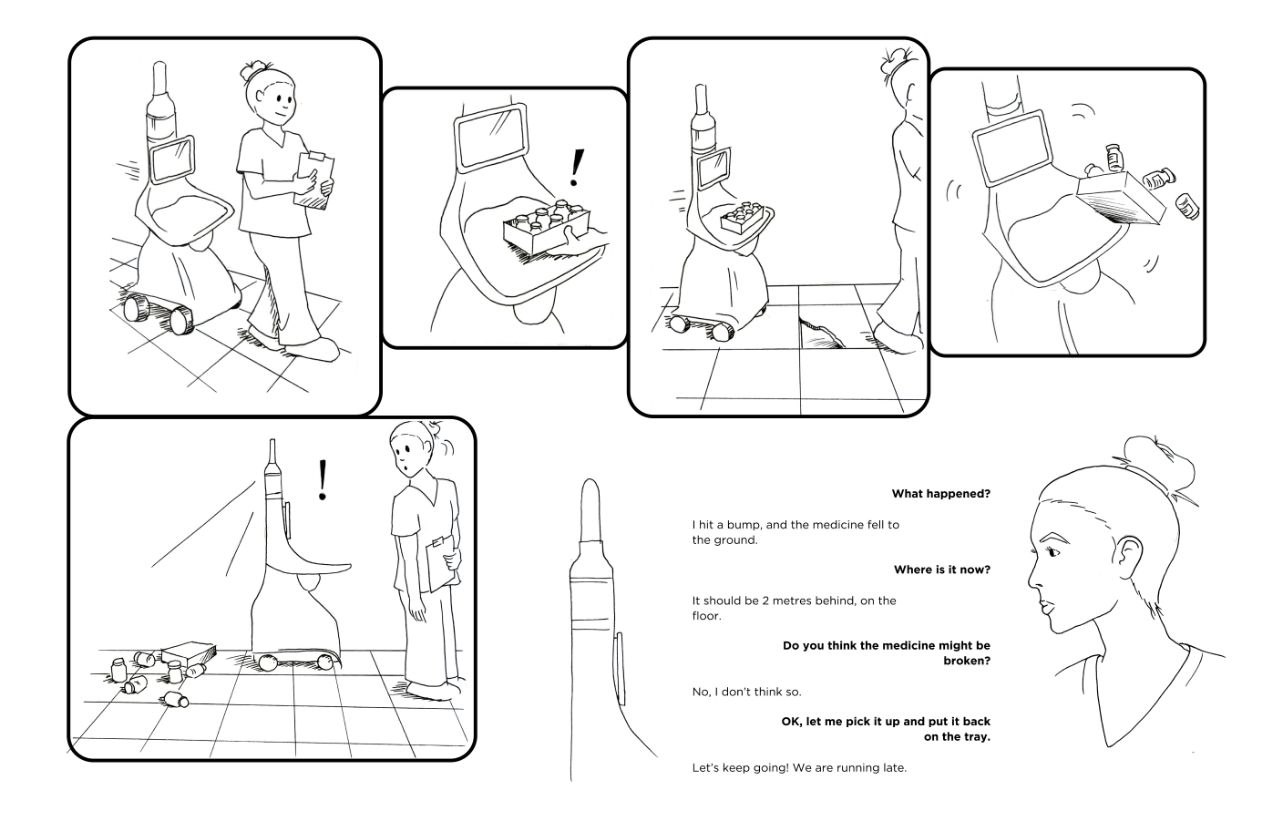}
    \caption{Perceptual anomaly and the search for a causal explanation}
    \label{fig:insight}
\end{figure}

This scenario provides the basic ingredients to trigger the search for a causal explanation and, more crucially, the use of the explanation to make structural changes in the robot's internal model, providing a potential way for learning a new concept, a new perceptual routine, or a new behaviour. Additionally, this causal explanation could be transformed into a natural language explanation for the surprised human that is leading the way. To be effective, this transformation should take into account the compatibility between models discussed earlier.    

\begin{figure}[t]
    \centering
    \includegraphics[width=1\linewidth]{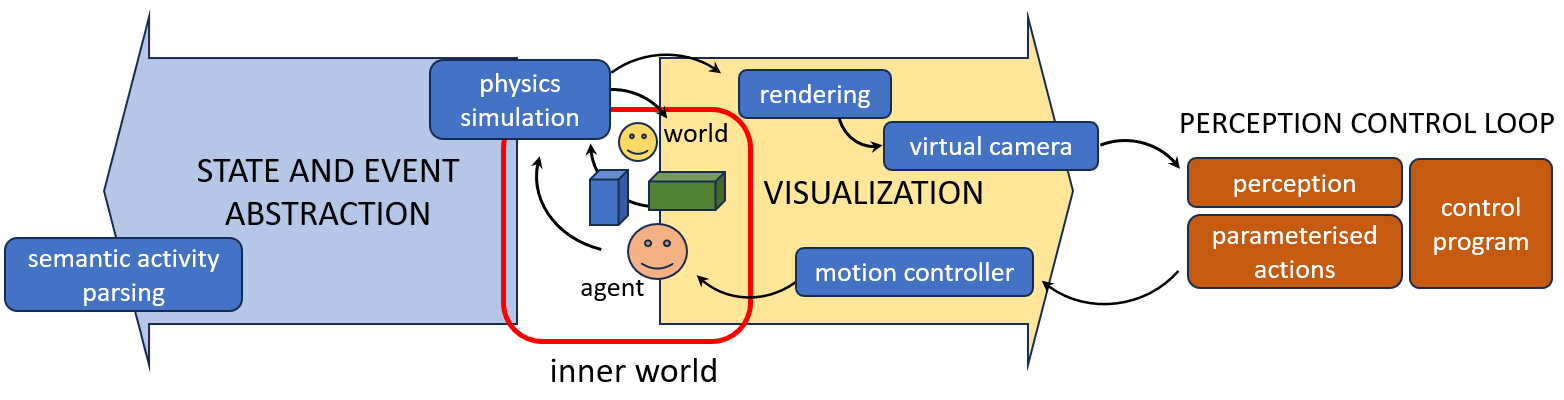}
    \caption{KnowRobSYM software architecture (see \citet{haidu:2018} for details)}
    \label{fig:knowrobsym}
\end{figure}

As suggested in \citet{Bustos2023}, the self-explanation of perceptual anomalies can be approached using an internal simulator as part of the robot's cognitive architecture. The use of internal simulators has gained momentum in recent years. Several research groups are including them in their cognitive robot architectures to explore, among other things, the trade-off between potential benefits and software complexity \cite{trinidad:2023,Mania2021,Kenghagho2022,bass_smith_bonawitz_ullman_2021}. These simulators have been used to generate predictions, select anticipatory actions, solve occlusion problems in object or human tracking, apply basic physical laws to perceived objects to stabilise them, or interpret human movements by projecting their kinematics into the robot's own model. A new potential use of internal simulators is their role as hypothesis testers in the construction of contrastive causal explanations. Following the scenario presented in Figure \ref{fig:insight}, the internal simulator in the robot would be used in two different ways. First, it would run in synchrony with the robot's perception and control system to generate short-term predictions of the perceived world. This mode is exactly what is needed to detect the perceptual anomaly caused by the bump in the first place. Since the robot did not see the bump, its internal simulator did not reflect it, and its predictions differed from the real IMU and camera measurements. This difference triggers the search for a cause that explains the anomaly and the new situation where the object is no longer on the tray. The second way of using the internal simulator will become clear in what follows. The search process could be divided into the following steps:

\begin{enumerate}
    \item the conversion of the anomaly data to a formal representation
    \item the search for a set of potential hypothesis 
    \item the conversion of each hypothesis into an initial state of the inner simulator
    \item the replay of the duration of the event since the detection of the anomaly starting with each selected hypothesis, and possible sweeping its internal parameters (i.e. position of the bump and width, depth or height)
    \item the comparison in each run of the synthetic and real sensorial data (i.e IMU) to check that the timings are an approximate match
    \item the selection of the best hypothesis and conversion to a formal representation, which can be used to modify and extend the internal knowledge of the robot and to generate a natural language explanation. 
\end{enumerate}
As this introspective sequence begins, the internal simulator is used as a replay machine, feeding from an episodic memory to reproduce the past event, starting from the occurrence of the anomaly. The scheme resembles the narrative-enabled episodic memories (NEEMs) proposed by \citet{beetz:2018}. NEEMs are constructed with an experience and a narrative. The NEEM experience stores the highly detailed recording of an experience (postures, perceptions, etc.), while the NEEM narrative is described as a story, with a more abstract or symbolic content. In \cite{Bustos2023}'s proposal, episodic memory shares geometric and symbolic elements in the same graph. As in KnowRobSIM \cite{haidu:2018}, the geometric part is manipulated using a game engine-based system. The software architecture of the game engine-enabled knowledge processing is schematised in Figure \ref{fig:knowrobsym}. Briefly, the inner world performs a loop in which (1) the dynamic state of the agent and world is updated (e.g. with control signals sent to the joint motors of the robot), (2) the world state is updated considering the control inputs generated by the agent (physics are taken into account), and (3) visually rendering the updated world state.

Returning to the example in Figure \ref{fig:insight}, the search for a set of plausible hypotheses could be based on an existing ontology and semantic memory, an external query to a wider resource, or even a prompt to an internal Large Language Model (LLM) \cite{zhang:2023}. With these initial conditions, the simulator can be used online to test whether the new potential causes can generate the sequence of sensory events previously registered. If so, the robot could conclude that it has found a plausible causal explanation for what happened and use it to improve its future behaviour and to communicate with the human. These two subsequent actions would eventually generate a confirmation or refutation of the explanation, feeding this learning process.

As this imagined scenario attempts to show, combining detectable perceptual anomalies with appropriate processes that can construct causal explanations, based on the acquisition and testing of a set of contrastive hypotheses, could be a promising path towards a self-explaining robotic cognitive architecture. Although there are many open problems to be solved, this approach is based from the outset on methods that will contribute to self-explaining robots. 

\section{Discussion and future research work} \label{sec3}

\begin{figure}[th]
\centering
  \includegraphics[width=1\textwidth]{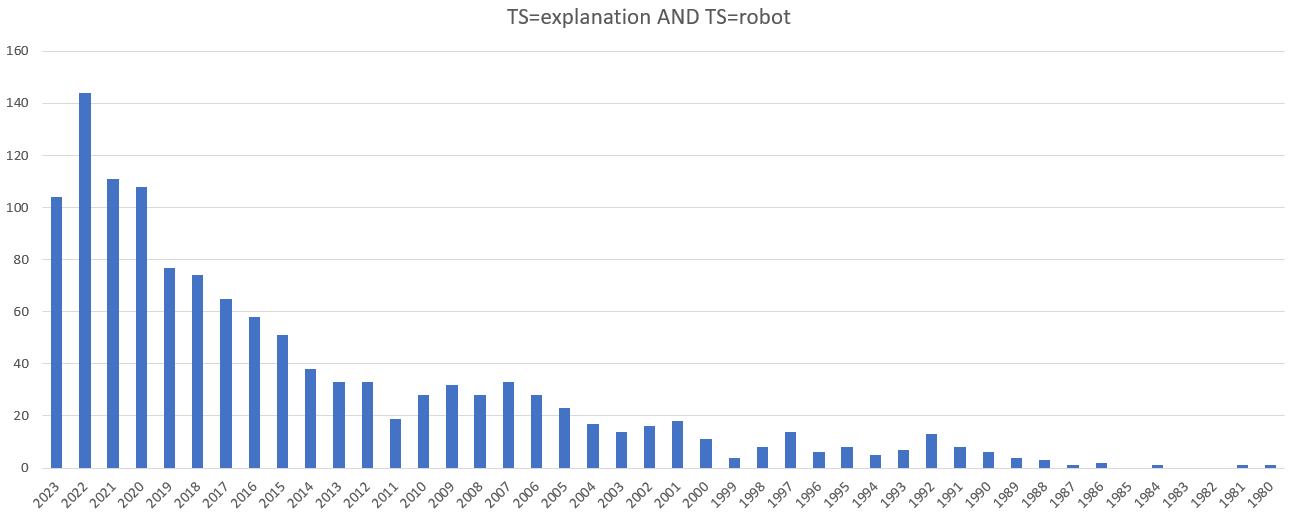}
\caption{Number of papers including the Topics Explanation and Robot. Web of Science Analyze filter: TS=(explanation) AND TS=(robot) (\url{https://clarivate.com/webofsciencegroup/solutions/web-of-science/}, Accessed on January 13, 2024)}
\label{fig:Papers}
\end{figure}

The last few decades have seen a true revolution in the perceptual capabilities that an autonomous robot can deploy. Algorithms that are already considered traditional have been practically replaced by approaches based on Deep Learning. Along with the evolution of algorithms, hardware and application scenarios have matured, so that today the robot can perceive the surrounding environment and build a robust and reliable semantic representation. Based on this perceptive capacity, and making use of algorithms based on the probabilistic approaches of the beginning of the century, autonomous robots are capable of solving certain tasks, such as navigation or localisation, in a robust manner. Equipped with all these hardware and software apparatus, autonomous robots are nowadays deployed in scenarios where they share space with people. It is not difficult to see them working as waiters in restaurants, or handling pallets in large intralogistics warehouses. However, deployment in scenarios where the nature of the tasks to be tackled, or the profile of the people with whom the environment is shared, means that the robot's behaviour is not always correctly interpreted by these people, forcing the robot to have to explain its behaviour. The field of work on the development of autonomous explainable robots is relatively young but has been gaining momentum in recent years. As Figure \ref{fig:Papers} shows, the number of publications considering the topics Explanation and Robot has grown exponentially since the beginning of the century (data are taken from the Web of Science). Although the data for 2023 is incomplete (the data capture date in the figure is January 2024), in recent years it has exceeded more than 100 publications per year, whereas in 2000 it was less than 20. 

Efforts in explanation generation have focused primarily on environments where robots and humans work together to solve a given task. In these cases, it is even possible for both actors to internalise the same causal model, which facilitates the detection of disparities. These disparities or inconsistencies will be the basis for the generation of explanations by the robot, always aimed at ensuring that the joint plan runs satisfactorily. In this same direction, the generation of explanations is linked to the existence of this plan, which the robot knows or learns, and which constitutes the reference that the robot seeks to be fulfilled. Adding constraints to the planning problem, with questions aimed for example at asking the robot to explain the reasons why one activity was executed instead of another, is framed as a restricted planning problem, which can be solved as the original unconstrained one and with a similar time.

Future work will need to address different issues. As discussed above, when task execution is shared between robots and humans, disparities arise from the side-by-side comparison of plans. This scheme does not consider the plans' causal structure to generate the explanations, nor does it go into details that might require abstracting in some way from the predicates, variables and/or actions. It is also relevant to consider the problem of unsolvability, i.e. the explanation of solutions or problems that complicate or impede the resolution of the task. In this sense, the use of solutions that consider geometric emulation (including, for example, physical models) in order to establish causal relationships at the level of the evolution of the context and not only of the action, can be helpful. 
An alternative approach uses PDDL to express planning problems and has important advantages, but it does not have expressiveness for considering constraints on inclusion, exclusion, or ordering actions, nor does it allow more complex constraints on the structure of the plan or on how something is achieved to be easily expressed (e.g., "Why did you use action A rather than action B for achieving P?"). To deal with this issue, instead of expressing constraints over the state trajectory of the plan (as the Linear Temporal Logic over finite traces \cite{giacomo:2014} or the PDDL3 \cite{gerevini:2009} suggest), plan constraints can be expressed over trajectories of actions \cite{bonassi:2022}. This will allow to improve performance by exploiting knowledge expressed by action constraints \cite{krarup:2021}. Additionally, there are less popular solutions that will also need to be explored.  One example is Answer Set Programming (ASP), designed for knowledge-intensive reasoning, but also used in task planning problems. In the work by \citet{jiang:2019}, ASP-based planners outperformed PDDL-based planners when the number of objects and tasks was large, and thus reasoning about complex precondition relationships and action effects was relevant.

The problem of explanation generation can also be approached from a slightly different angle: the generation of explanations triggered by perceptual anomalies and the translation of these explanations into a human-understandable format. These anomalies are likely to occur frequently in the daily operation of a social robot. Internal simulators can be an effective way to test contrastive hypotheses online, but several problems remain to be solved. One of the most challenging is the online transfer between the numerical representations needed to run simulations and the symbolic ones needed to search and extend the robot's internal knowledge. This coming and going of the transformations required to build a new causal explanation is, from the outset, a promising place to pursue research in the creation of new concepts.

\section{Conclusions} \label{sec4}

This article describes the general framework of the generation of explanations by an autonomous robot, starting from the basic concept of its internalization of a more or less complex scheme of cause-effect relationships. The review of the state of the art allows us to assess how, in recent years and after the great evolutionary leap experienced in the perceptual abilities of these robots, the integration of causality has begun to be considered as one of the basic pillars of the creation of more acceptable robots that can be considered by people as credible, safe and reliable. Specifically, this causality serves as the basis for the robot to generate explanations, either limited to a temporal sequence that is being traversed (with a series of specific actions, already executed or to be executed in the future) or open to the analysis of temporal sequences not initially considered (in which there are actions that could have been executed but have not been, or in which perceptions have been disregarded, which may be the cause of unforeseen results).

\bibliographystyle{plainnat}
\bibliography{mybibfile}

\end{document}